\documentclass{sig-alternate-05-2015}

\begin{document}

\setcopyright{acmcopyright}

\doi{http://dx.doi.org/10.1145/3006299.3006310}

\isbn{978-1-4503-4617-7/16/12}


\acmPrice{\$15.00}

%
\conferenceinfo{BDCAT'16,}{December 06-09, 2016, Shanghai, China}
\CopyrightYear{2016} 

\title{Neighborhood Features Help Detecting Non-Technical Losses in Big Data Sets}

%
%
%
%
%

\numberofauthors{7} 
%
\author{
%
%
\alignauthor
Patrick Glauner\\
       \affaddr{Interdisciplinary Centre for Security, Reliability and Trust, University of Luxembourg}\\
       \affaddr{4, rue Alphonse Weicker}\\
       \affaddr{2721 Luxembourg City, Luxembourg}\\
       \email{patrick.glauner@uni.lu}
\alignauthor
Jorge Augusto Meira\\
       \affaddr{Interdisciplinary Centre for Security, Reliability and Trust, University of Luxembourg}\\
       \affaddr{4, rue Alphonse Weicker}\\
       \affaddr{2721 Luxembourg City, Luxembourg}\\
       \email{jorge.meira@uni.lu}
\alignauthor
Lautaro Dolberg\\
       \affaddr{CHOICE Technologies Holding S\`arl}\\
       \affaddr{2-4, rue Eug\`ene Ruppert}\\
       \affaddr{2453 Luxembourg City, Luxembourg}\\
       \email{lautaro.dolberg\\@choiceholding.com}
\and  
\alignauthor
Radu State\\
       \affaddr{Interdisciplinary Centre for Security, Reliability and Trust, University of Luxembourg}\\
       \affaddr{4, rue Alphonse Weicker}\\
       \affaddr{2721 Luxembourg City, Luxembourg}\\
       \email{radu.state@uni.lu}
\alignauthor
Franck Bettinger\\
       \affaddr{CHOICE Technologies Holding S\`arl}\\
       \affaddr{2-4, rue Eug\`ene Ruppert}\\
       \affaddr{2453 Luxembourg City, Luxembourg}\\
       \email{franck.bettinger\\@choiceholding.com}
       \alignauthor
Yves Rangoni\\
       \affaddr{CHOICE Technologies Holding S\`arl}\\
       \affaddr{2-4, rue Eug\`ene Ruppert}\\
       \affaddr{2453 Luxembourg City, Luxembourg}\\
       \email{yves.rangoni\\@choiceholding.com}
}
\additionalauthors{Additional author: Diogo Duarte (CHOICE Technologies Holding S\`arl,
email: {\texttt{diogo.duarte@choiceholding.com}}).}
\date{5 October 2016}

\maketitle
\begin{abstract}
Electricity theft occurs around the world in both developed and developing countries and may range up to 40\% of the total electricity distributed. More generally, electricity theft belongs to non-technical losses (NTL), which occur during the distribution of electricity in power grids.
In this paper, we build features from the neighborhood of customers. We first split the area in which the customers are located into grids of different sizes. For each grid cell we then compute the proportion of inspected customers and the proportion of NTL found among the inspected customers. We then analyze the distributions of features generated and show why they are useful to predict NTL.
In addition, we compute features from the consumption time series of customers. We also use master data features of customers, such as their customer class and voltage of their connection.
We compute these features for a Big Data base of 31M meter readings, 700K customers and 400K inspection results.
We then use these features to train four machine learning algorithms that are particularly suitable for Big Data sets because of their parallelizable structure: logistic regression, k-nearest neighbors, linear support vector machine and random forest.
Using the neighborhood features instead of only analyzing the time series has resulted in appreciable results for Big Data sets for varying NTL proportions of 1\%-90\%. This work can therefore be deployed to a wide range of different regions.
\end{abstract}

%
%


%
%

%
%



\section{Introduction}
Modern life would be unimaginable without reliable availability of electricity. Electricity is generated by different infrastructure such as power plants, wind farms or solar cells. It is then distributed to customers through electrical power grids. There are frequently appearing losses, of which electricity theft is most predominantly known to the public. However, losses can be classified more accurately into technical and non-technical losses. Technical losses are naturally caused due to power dissipation, in particular by internal electrical resistance of the wires. The focus of this paper is on non-technical losses (NTL), which appear during distribution and include electricity theft.

There are many forms of electricity theft such as meter tampering, bypassing meters or arranged false meter readings, for example by manipulating interfaces or bribing meter readers \cite{chauhan2013non}. Other forms of NTL include faulty or broken meters, un-metered supply and human or technical errors in meter readings, processing and billing \cite{smith2004electricity}. This paper therefore considers not only electricity theft, but NTL as a whole.

\begin{figure}[h!]
\centering
\includegraphics[width=0.475\textwidth]{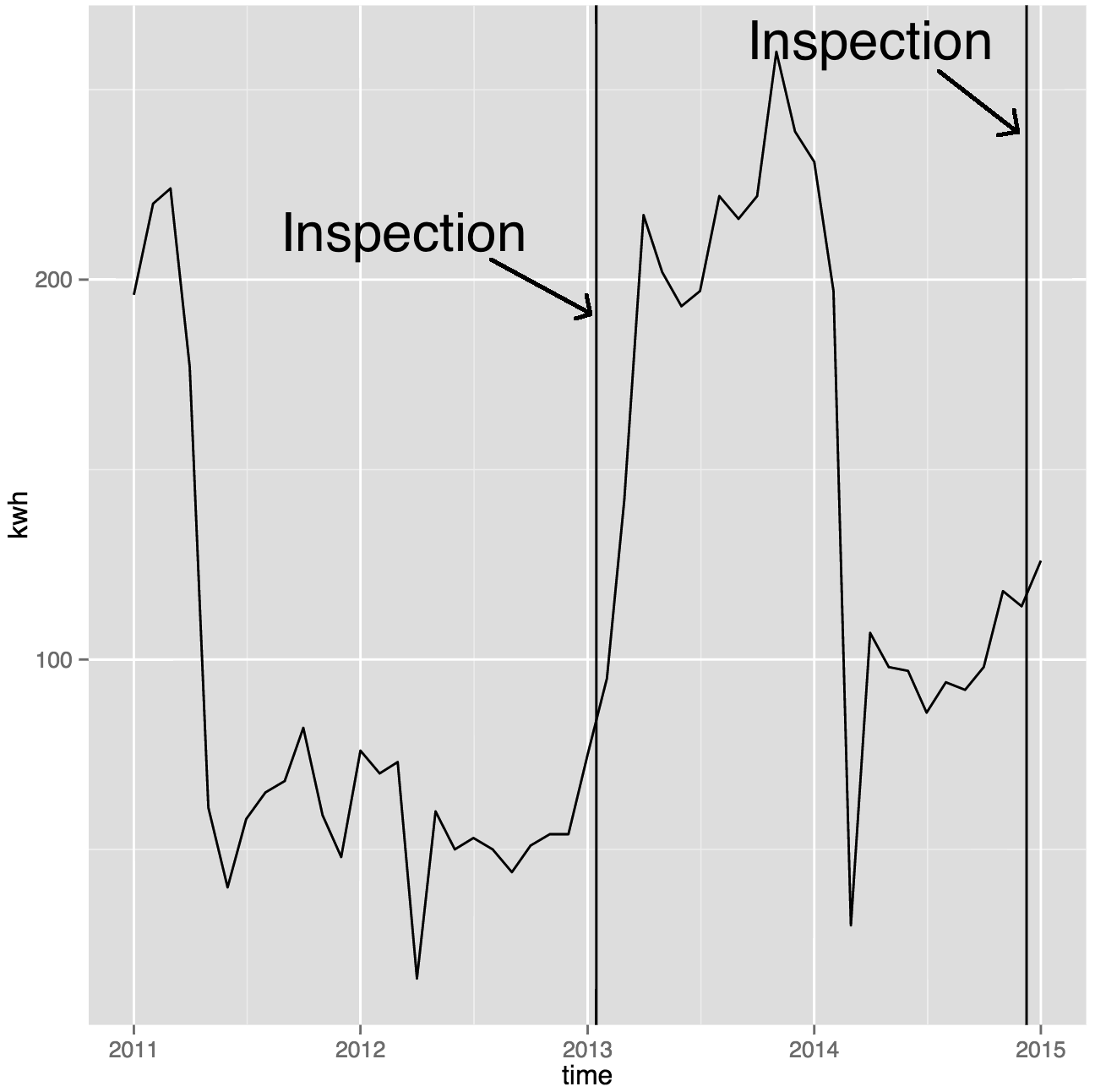}
\caption[XXX]{Two assumed occurrences of NTL due to significant consumption drop followed by inspections (visualized by a vertical bar).}
\label{fig:fraud_example}
\end{figure}

An example of a consumption time series of a customer with monthly meter readings is depicted in Figure~\ref{fig:fraud_example}. In the beginning of 2011, the consumption significantly decreased to about a fifth and remained at this level in the course of 2011. Based on this pattern, an inspection was carried out in the beginning of 2013, which detected a NTL, more concretely electricity theft. This manipulation of the infrastructure was reverted and the electricity consumption resumed to the previous level. One year later, the electricity consumption dropped again to about a third, which led to another inspection a few months later.

NTLs are of significant harm to economies and effects include loss of revenue and profit, decrease of the stability and reliability of power grids. They are reported to range up to 40\% of the total electricity distributed in countries such as Brazil, India, Malaysia or Lebanon \cite{depuru2013high}. They are also of relevance in developed countries, for example estimates of NTLs in the UK and US range from US\$ 1-6 billion \cite{alam2004power}. Carrying out physical inspections of customers for NTL is costly. Therefore, NTL predictions have to be accurate.

We have previously identified the main open challenges in order to advance NTL detection \cite{glauner2016challenge}:
\begin{itemize}
\item Class imbalance and evaluation metric
\item Feature description
\item Incorrect inspection results
\item Biased inspection results
\item Scalability
\item Comparison of different methods
\end{itemize}

Imbalanced classes in a data set describe the fact that it contains an unequal amount of labels per class. Accurate evaluation metrics need to take this property into account.
Feature description is a long-standing challenge in machine learning because learning algorithms often do not work on the raw data and need to be trained on features computed from the raw data. The set of inspected customers is a sample of all customers. This sample does not represent the overall population of customers as previous inspections have focused on certain area. Furthermore, some inspection results reported are incorrect as technicians may have been threatened or bribed by fraudsters.

The main idea of this paper is to build scalable models that use neighborhood-based engineered features from imbalanced Big Data sets. By using information of the neighborhood, we can predict NTL better as there are geographic clusters of NTL among the customers. To the best of our knowledge, we are not aware of any previously published research that addressed this topic.

The rest of this paper is organized as follows. Section \ref{chapter:review} provides a literature review of NTL detection. Section \ref{chapter:ntl} describes the data set and the main contribution, which are features that include information about the neighborhood. We analyze their statistical properties and show why they are useful for NTL detection. Furthermore, we describe the different proposed NTL detection models and explain why they are particularly scalable to Big Data sets. Section \ref{chapter:eval} presents experimental results and a comparison of the models on the data for different NTL proportions in the data. Section \ref{chapter:end} summarizes this work and provides an outreach on future work.

\section{Related Work}
\label{chapter:review}
NTL detection can be treated as an anomaly or fraud detection problem. Comprehensive surveys of the field of NTL detection are provided in \cite{chauhan2013non}, \cite{glauner2016challenge} and \cite{kazerooni2014literature}. Surveys of how an advanced metering infrastructure can be manipulated, are provided in \cite{jiang2014energy} and \cite{mclaughlin2009energy}.
Most NTL detection research apply artificial intelligence methods to it. In particular, the methods used fall into two categories: expert systems and machine learning. Expert systems incorporate hand-crafted rules in order to make decisions. In contrast, machine learning methods learn models from data without being explicitly programmed.

One method to detect NTL is to derive features from the customer consumption time series, such as in \cite{angelos2011detection}: average consumption, maximum consumption, standard deviation, number of inspections and average consumption of the residential neighborhood. These features are then grouped into c classes using fuzzy c-means clustering. Next, customers are classified into NTL or no NTL using the fuzzy memberships. An average precision of 0.745 is achieved on the test set.

Daily average consumption features of the last 25 months are used in \cite{nagi2010nontechnical} for less than 400 out of a highly imbalanced data set of 260K customers. These features are then used in a support vector machine (SVM) with a Gaussian kernel for NTL prediction, for which a test recall of 0.53 is achieved.

The class imbalance problem has been addressed in \cite{di2012improving}. In that paper, an ensemble of two SVMs, an optimum-path forest and a decision tree is applied to 300 test data. While the class imbalance problem is addressed, the degree of imbalance of the 1.5K training examples is not reported.

The consumption profiles of 5K Brazilian industrial customer profiles are analyzed in \cite{ramos2012identification}. Each customer profile contains 10 features including the demand billed, maximum demand, installed power, etc. A SVM and k-nearest neighbors perform similarly well with test accuracies of 0.962. Both outperform a neural network, which achieves a test accuracy of 0.945.

In our previous research in \cite{glauner2016large} we have particularly addressed the class imbalance of NTL detection and how to assess models in such an environment. We have compared Boolean and fuzzy expert systems to a support vector machine trained on time series features of NTL proportion samples ranging from 0.1\% to 90\% of 700K Brazilian customers. In order to assess the models, we have proposed to use the area under the receiver-operating characteristic curve (AUC).

We have shown that only analyzing the time series leads to limited results and that further data must be taken into account, too. We believe that the neighborhood of customers contains information about whether a customer may cause a NTL or not. This has not adequately been addressed in previous research. Furthermore, many models reported in the literature do not scale to Big Data sets and from our perspective it is necessary to take this into account in order to deploy the models to a real environment.

\section{NTL Detection based on Neighborhood Features}
\label{chapter:ntl}

\subsection{Data}
The data set used for experiments in this paper is from an electricity provider in Brazil. It contains 31M monthly meter readings from January 2011 to January 2015 of 700K customers. Each meter reading contains the consumption in kWh and date of the reading. For each customer, the master data includes, but is not limited to, the location, the customer class, the voltage of the connection, the number of wires going into the building and the contract status. Also, the data set includes 400K inspection results: if a NTL was found, the type of NTL and notes written by the technician. This is the same database used in our previous research \cite{glauner2016large}.

About one third of the inspections found a NTL. However, the models of this paper must also work in other regions which have different NTL proportions. Therefore, 14 samples each having 100K inspects results are generated with the following NTL proportions: 1\%, 2\%, 3\%, 4\%, 5\%, 10\%, 20\%, 30\%, 40\%, 50\%, 60\%, 70\%, 80\% and 90\%.

\subsection{Features}
For each proportion sample, the following three types of features are used per customer $m$: neighborhood information, daily average consumption and categorial master data of the customer. Also, a binary target vector $T$ is created in which element $T^{(m)}$ is the most recent inspection result for customer $m$ in the respective period of time. NTLs are encoded by 1 if they are detected and 0 if not.

\subsubsection{Neighborhood}
Certain areas are more likely to cause NTLs than others. Therefore, features based on the neighborhood are interesting in order to improve predictions. 
The data includes invalid coordinates of customers, such as coordinates in the ocean. For this, all customers outside a deviation from the mean coordinates are removed. We empirically found that removing the 1K customers that are not within five standard deviations from the mean coordinates worked the best.
The bounding box around the remaining valid coordinates is about 200 km along the longitude and about 500 km along the latitude. Therefore, the bounding box has an area of approximately 100,000 km$^2$. This bounding box is split into a grid along the longitude and latitude.

In each cell$_{ij}$, the proportion of inspected customers and the proportion of NTL found among the inspected customers are computed:
\begin{align}
\operatorname{inspected\_ratio}_{ij} = \frac{\operatorname{\#inspected}_{ij}}{\operatorname{\#customers}_{ij}},
\end{align}
\begin{align}
\operatorname{NTL\_ratio}_{ij} = \frac{\operatorname{\#NTL}_{ij}}{\operatorname{\#inspected}_{ij}}.
\end{align}

\begin{figure}[h!]
\centering
\includegraphics[width=0.3\textwidth]{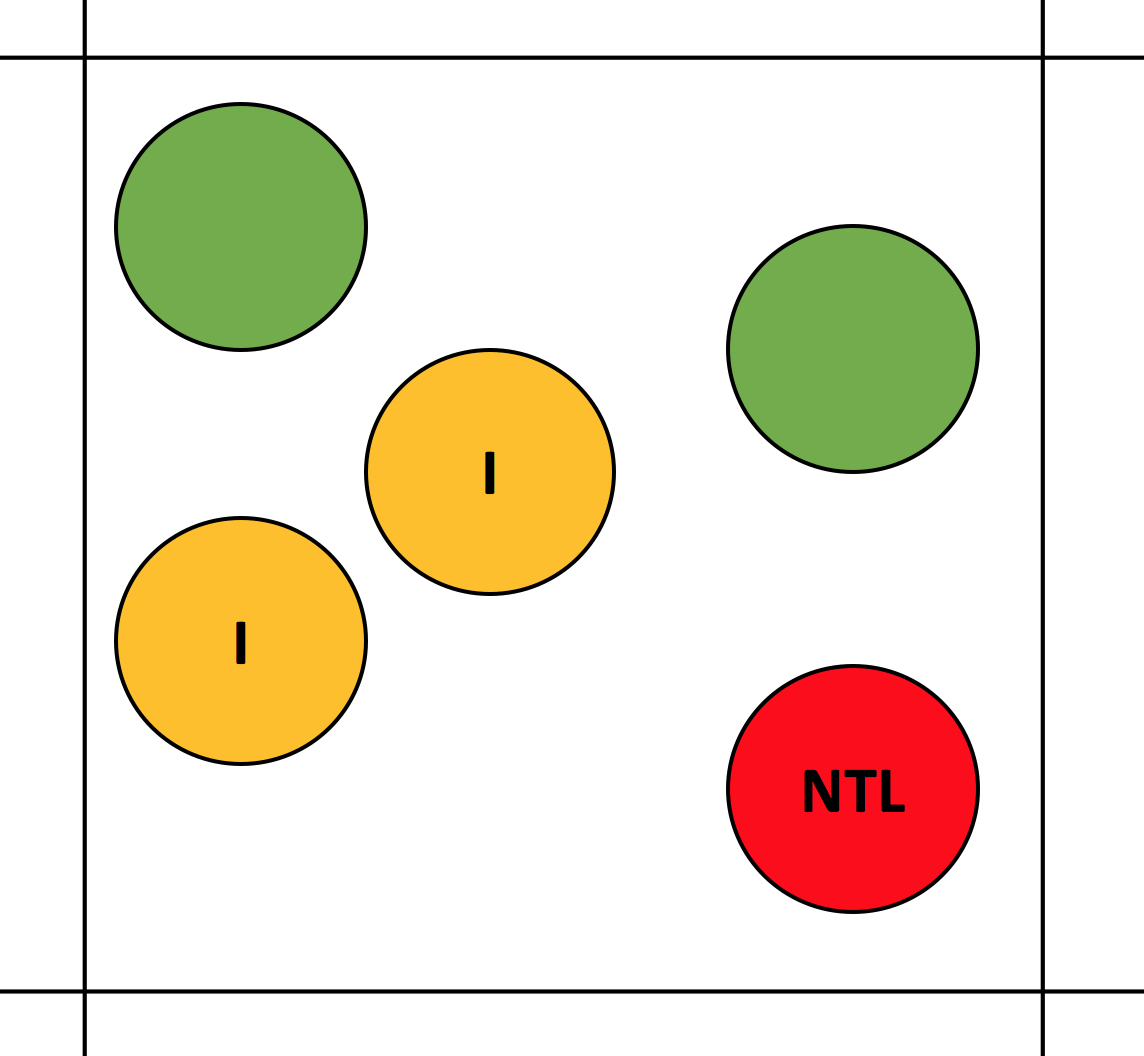}
\caption[XXX]{Example cell with 5 customers, 3 out of 5 were inspected (I) and 1 out of 3 inspected customers caused a NTL.}
\label{fig:grid_example}
\end{figure}

An example cell is provided in Figure~\ref{fig:grid_example}.

The grid sizes used are 50, 100, 200 and 400 cells along the longitude and latitude, respectively. For each grid size, both features are assigned to each customer registered in the respective cell. The area per cell is depicted in Table~\ref{table:cellarea} for each grid size.

\begin{table}[h!]
\renewcommand{\arraystretch}{1.3}
\caption{Area Per Cell for All Grid Sizes}
\begin{minipage} {0.5\textwidth}
\label{table:cellarea}
\begin{center}
\centering
\begin{tabular}{|c|c|}
\hline
Grid size & Area per cell [km$^2$] \\
\hline
$50\times 50$ & 40 \\
\hline
$100\times 100$ & 10 \\
\hline
$200\times 200$ & 2.5 \\
\hline
$400\times 400$ & 0.625 \\
\hline
\end{tabular}
\end{center}
The total area of the bounding box around the customers is approximately 500 km $\times$ 200 km = 100,000 km$^2$.
\end{minipage}
\end{table}

\begin{figure}[h!]
\centering
\includegraphics[width=0.475\textwidth]{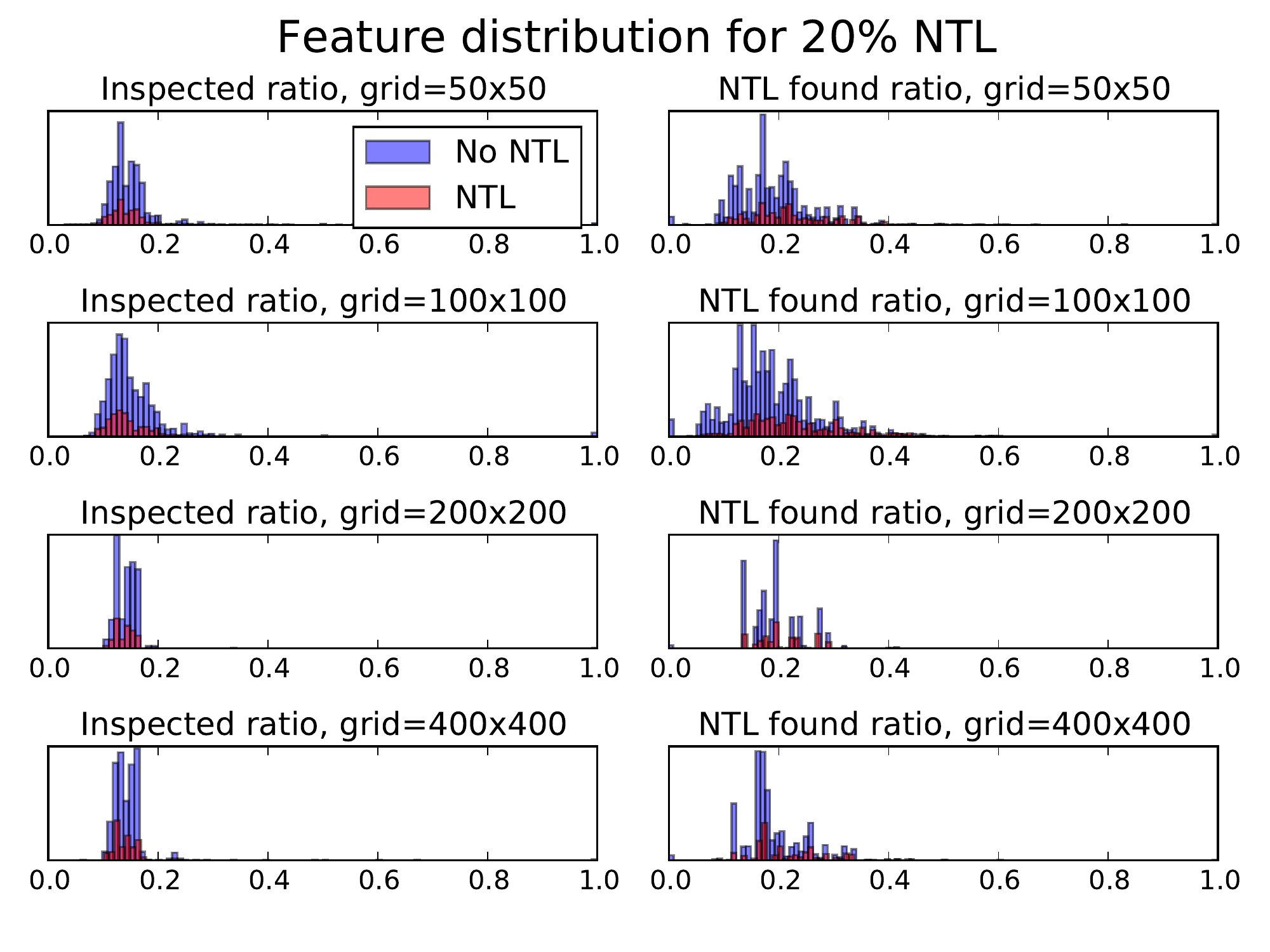}
\caption[XXX]{Distributions of both neighborhood features for varying grid sizes for 20\% NTL.}
\label{fig:dist20}
\end{figure}

\begin{figure}[h!]
\centering
\includegraphics[width=0.475\textwidth]{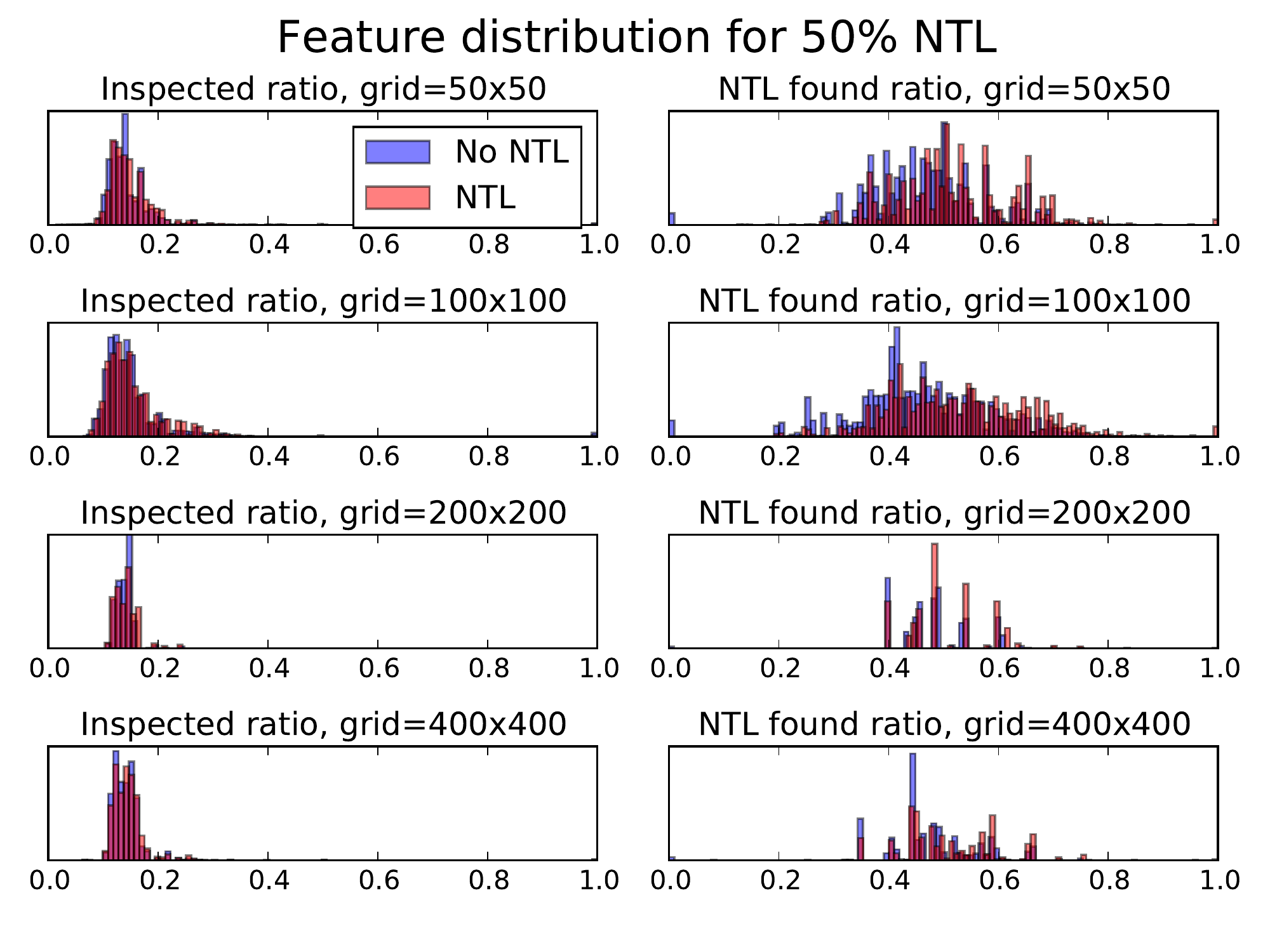}
\caption[XXX]{Distributions of both neighborhood features for varying grid sizes for 50\% NTL. Example: the red NTL peak around 0.5 in the NTL found ratio, grid=200x200 plot represents a type of favela neighborhood, in which every second customer causes a NTL.}
\label{fig:dist50}
\end{figure}

As four grid sizes are used, a total of $4\times 2 = 8$ neighborhood features are computed per customer.
For both classes, the distributions of the values of both features for these four grid sizes are depicted in Figure~\ref{fig:dist20} for a NTL proportion of 20\% and in Figure~\ref{fig:dist50} for a balanced NTL proportion.

The distributions of both neighborhood features represent the prior distributions of a Bayesian approach.
However, none of the distributions is Gaussian, and it is therefore interesting to study how their properties change for varying NTL proportions of the data set and how they allow to separate between no NTL found and NTL found.

\begin{figure}[h!]
\centering
\includegraphics[width=0.475\textwidth]{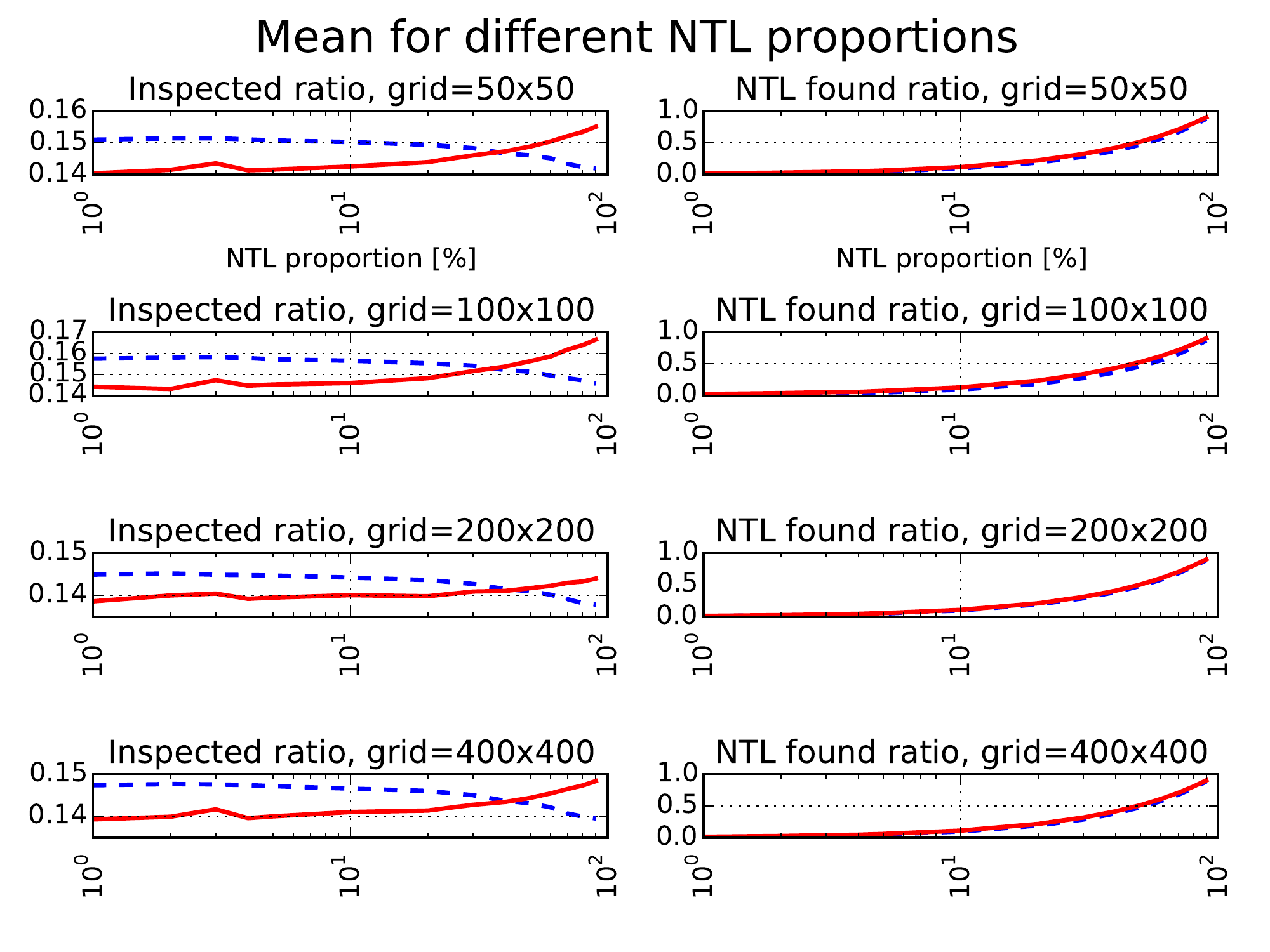}
\caption[XXX]{Mean of each feature distribution for different NTL proportions. Legend: the blue dashed curve represents the no NTL class and the red solid curve represents the NTL class.}
\label{fig:nmean}
\end{figure}

The mean of each feature distribution is depicted in Figure~\ref{fig:nmean}.
The means of the inspected ratio distributions are expected to be around 0.14 because there are 700K customers and each NTL proportion file contains 100K inspections. However, the means slightly decrease for greater NTL proportions for customers for which no NTL was found and slightly increase for customers for which NTL was found. We have not found any cause of this in our experiments, however, we believe that this is caused by the sampling of the data. However, this helps to separate both classes.
The means of the NTL found ratio distributions are approximately the NTL proportion as expected. 
For all grid sizes, the distributions of means are approximately the same for the inspected ratio and NTL found ratio features, respectively.

\begin{figure}[h!]
\centering
\includegraphics[width=0.475\textwidth]{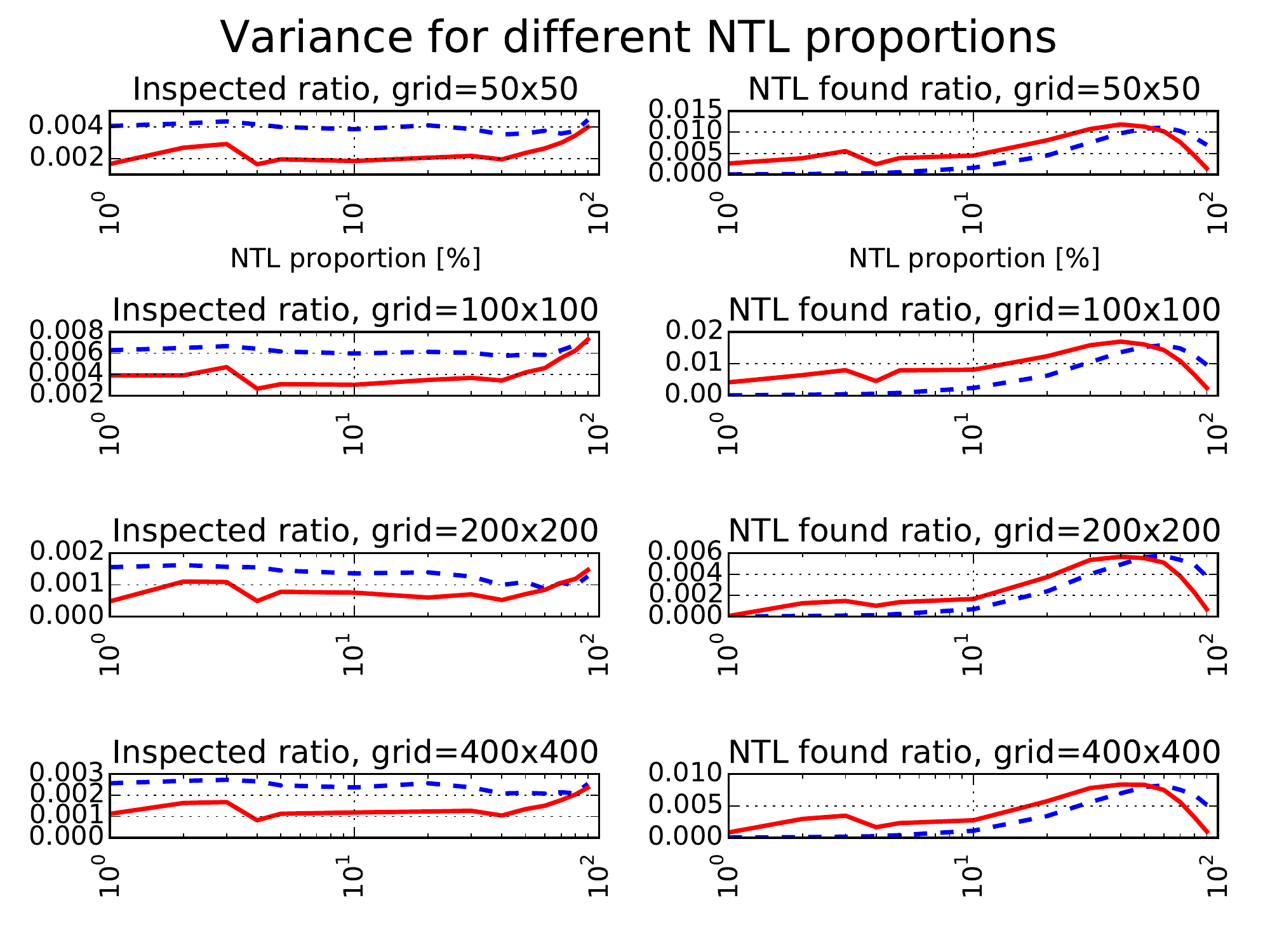}
\caption[XXX]{Variance of each feature distribution for different NTL proportions. Legend: the blue dashed curve represents the no NTL class and the red solid curve represents the NTL class.}
\label{fig:nvariance}
\end{figure}

The variance of each feature distribution is depicted in Figure~\ref{fig:nvariance}.
For the inspected ratio feature, we see that the variance is lower for the customers for which NTL was found than for those for which no NTL was found. There is only an exception for the grid size of 100 for NTL proportions $> 70\%$.
The variance of the NTL found ratio feature is greater for the customers for which NTL was found than those for which no NTL was found for NTL proportions $< 50\%$ and then flips around 50\% for all grid sizes.
This demonstrates an inverse relationship between the distributions of variances of both features for NTL proportions $< 50\%$. For both features, the variances are in different ranges for each grid sizes, which helps to separate between both classes.

\begin{figure}[h!]
\centering
\includegraphics[width=0.475\textwidth]{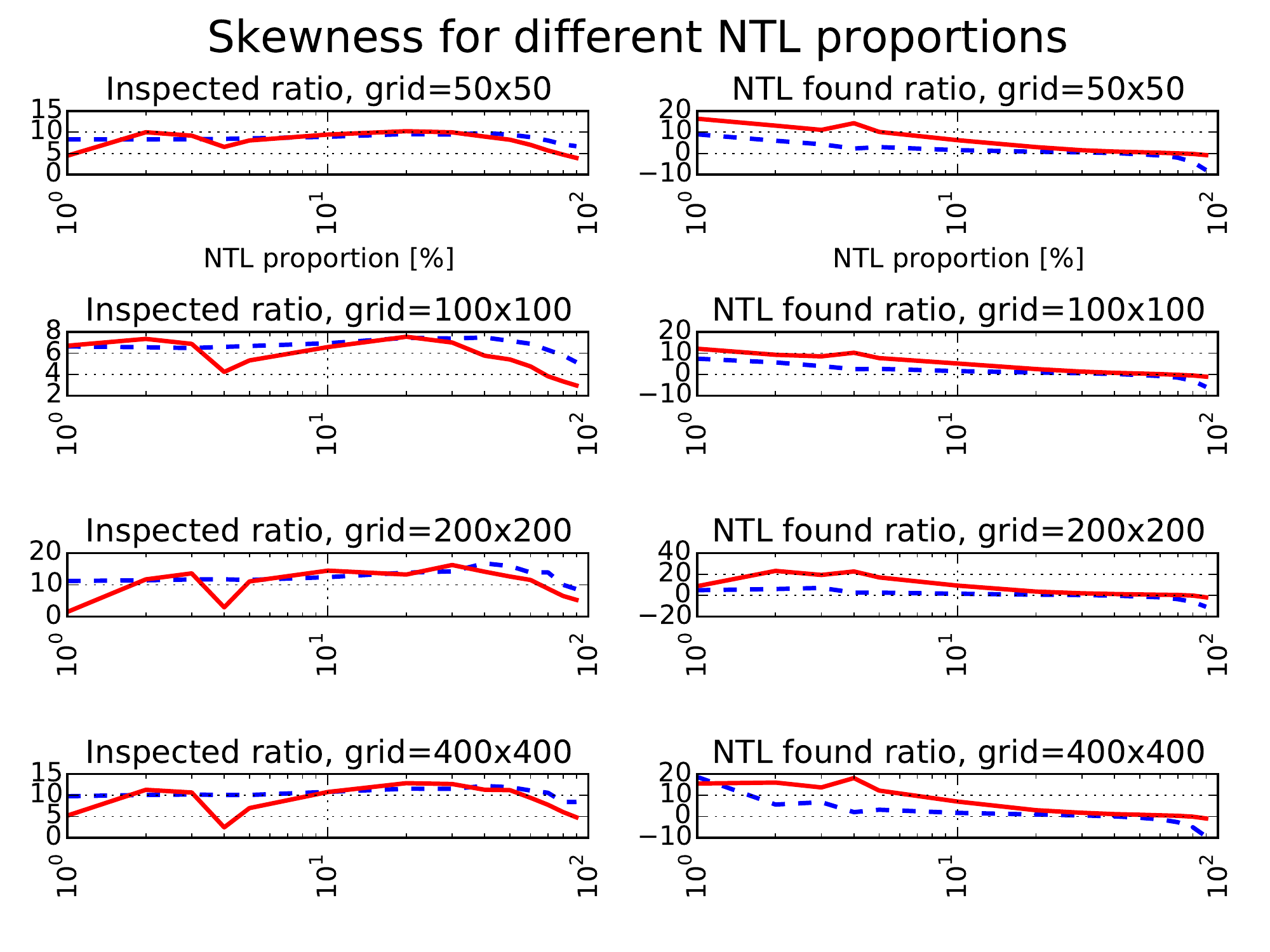}
\caption[XXX]{Skewness of each feature distribution for different NTL proportions. Legend: the blue dashed curve represents the no NTL class and the red solid curve represents the NTL class.}
\label{fig:nskewness}
\end{figure}

Skewness is the extent to which the data are not symmetrical \cite{decarlo1997meaning}. It is the third standardized moment, defined as:
\begin{align}
    \gamma_1 = \operatorname{E}\left[\left(\frac{X-\mu}{\sigma}\right)^3 \right]
                 = \frac{\mu_3}{\sigma^3}
             = \frac{\operatorname{E}\left[(X-\mu)^3\right]}{\ \ \ ( \operatorname{E}\left[ (X-\mu)^2 \right] )^{3/2}},
\end{align}
where $\mu_3$ is the third central moment, $\mu$ is the mean, $\sigma$ is the standard deviation and $\operatorname{E}$ is the expectation operator.
Positively skewed data have a tail that points to the right. In contrast, negatively skewed data have a tail that points to the left.
The skewness of each feature distribution is depicted in Figure~\ref{fig:nskewness}. 

All inspected ratio distributions are positively skewed. This skewness means that there are more grid cells with very high inspected ratios than cells with very low inspection ratios. There is no significant difference between both classes for most NTL proportions and therefore this property does not help much to separate between both.
All NTL found ratio distributions are positively skewed for NTL proportions $\le 50\%$. For NTL proportions $> 50\%$, the distributions are negatively skewed for the no NTL class.
The change of sign in the skewness distributions for samples with low NTL proportions shows the existence of clusters of low NTLs of different sizes. The skewness of this feature is generally greater for the NTL class than the no NTL class for all grid sizes, which allows to separate both classes better.

\begin{figure}[h!]
\centering
\includegraphics[width=0.475\textwidth]{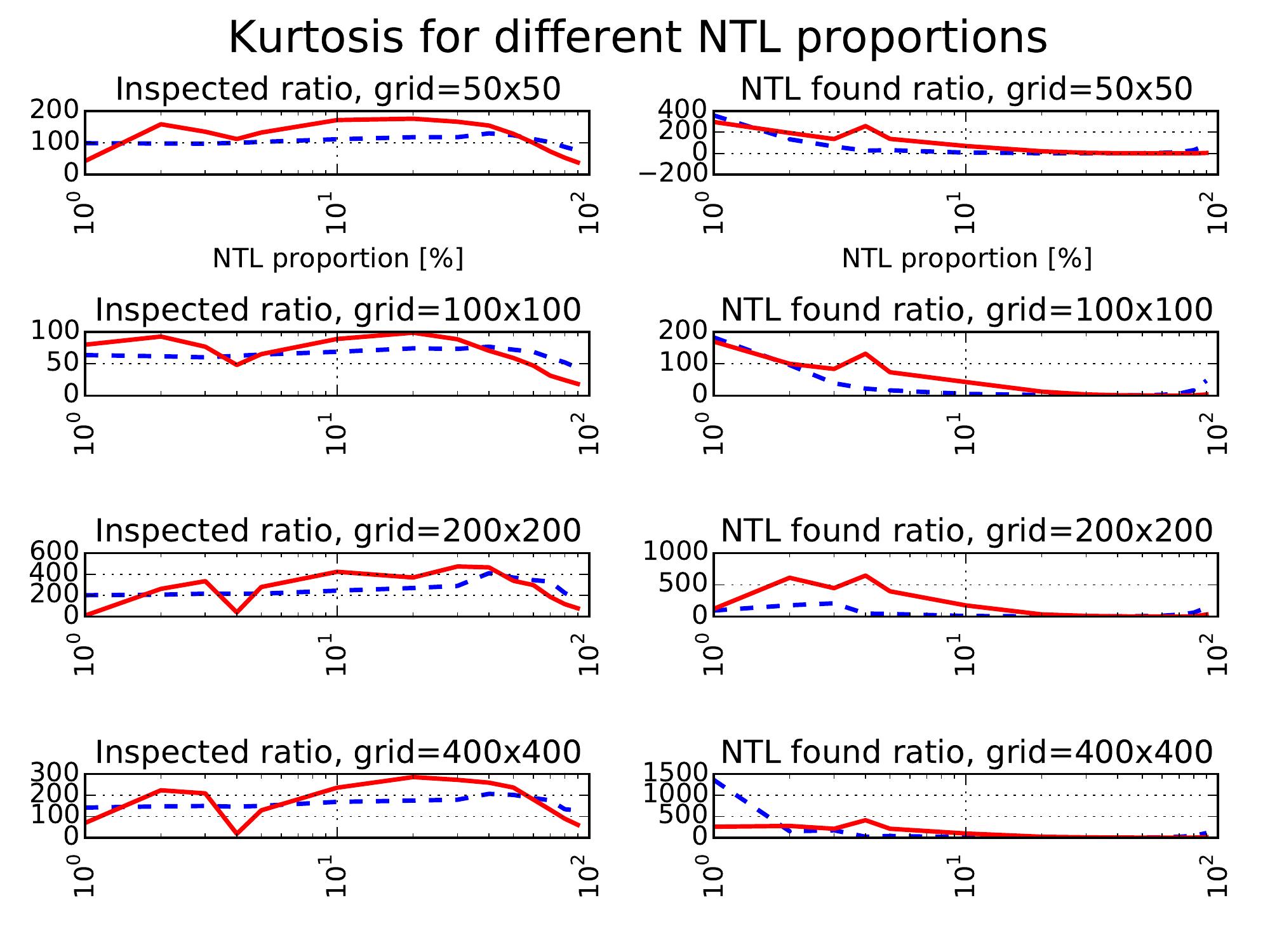}
\caption[XXX]{Kurtosis of each feature distribution for different NTL proportions. Legend: the blue dashed curve represents the no NTL class and the red solid curve represents the NTL class.}
\label{fig:nkurtosis}
\end{figure}

Kurtosis indicates how the peak and tails of a distribution differ from the normal distribution \cite{decarlo1997meaning}. It is the fourth standardized moment, defined as:
\begin{align}
\operatorname{Kurt}[X] = \frac{\mu_4}{\sigma^4} = \frac{\operatorname{E}[(X-{\mu})^4]}{(\operatorname{E}[(X-{\mu})^2])^2} -3,
\end{align}
where $\mu_4$ is the fourth moment about the mean and $\sigma$ is the standard deviation.
A distribution with a positive kurtosis value has heavier tails and a sharper peak than the normal distribution. In contrast, a distribution with a negative kurtosis value indicates that the distribution has lighter tails and a flatter peak than the normal distribution.
The kurtosis of each feature distribution is depicted in Figure~\ref{fig:nkurtosis}. 

The kurtosis values of all distributions of both features are positive and therefore have sharper peaks than the normal distribution.
For the inspection ratio features, the kurtosis is greater for the NTL class for most NTL proportions, meaning these features are less Gaussian than for the NTL class, which helps to separate both classes.
The same applies to the NTL found ratio feature for NTL proportions $< 50\%$.

Overall, the plots of variance, skewness and kurtosis of both classes show that for both features the values of the distributions for the different grid sizes have different ranges. This is helpful in order to discriminate between NTL and no NTL.


\subsubsection{Daily Average Consumption}
A daily average consumption feature during month $d$ for customer $m$ in kWh is:
\begin{align}
x_d^{(m)} = \frac{L_d^{(m)}}{R^{(m)}_{d} - R^{(m)}_{d-1}}.
\end{align}
This feature is computed for $M$ customers $\{0, 1, ..., M - 1\}$ over the last $N$ months $\{0, 1, ..., N -1\}$.
$L_d^{(m)}$ is the consumption in kWh between the meter reading $R^{(m)}_d$ of month $d$ and the previous one $R^{(m)}_{d-1}$ in month $d-1$. $R^{(m)}_{d} - R^{(m)}_{d-1}$ is the number of days between both meter readings of customer $m$. These features are based on \cite{nagi2010nontechnical} and are also used in our previous research \cite{glauner2016large}.
For the experiments in Section~\ref{chapter:eval}, $N=12$ was experimentally determined to work the best.

\subsubsection{Categorial Master Data}
In addition, more information about the customer should be considered in the prediction. The categorial master data available for each customer is summarized in Table~\ref{table:masterdatafeatures}. Each feature is converted to one-hot coding. Therefore, there are $8 + 3 + 3 + 2 = 16$ binary features per customer.

\begin{table}[h!]
\renewcommand{\arraystretch}{1.3}
\caption{Available Master Data}
\label{table:masterdatafeatures}
\centering
\begin{tabular}{|c|c|}
\hline
Name & Possible values  \\
\hline
Class & Residential, commercial, industrial, \\
& public illumination, rural, \\
& public service, power \\
& generation infrastructure \\
\hline
Contract status & Active, suspended, inactive \\
\hline
Number of wires & 1, 2, 3 \\
\hline
Voltage & \textgreater 2.3kV, $\le$2.3kV \\
\hline
\end{tabular}
\end{table}

In order to reduce overfitting, only representative binary features are kept. These could be found using the principal component analysis (PCA). However, PCA is not able to handle noise in the data well. Since this real data set is noisy, PCA is not used for the reduction of the binary features.
Instead the dimensionality reduction approach is as follows: All features that are either one or zero in more than $p\times 100$\% of each proportion sample are removed. These binary features are Bernoulli random variables, and the variance of such variables is given by:
\begin{align}
\operatorname{Var}[X] = p(1-p).
\end{align}

For the experiments in Section~\ref{chapter:eval}, $p=0.9$ was experimentally determined to work the best.

\subsubsection{Final Feature Set}
\label{chapter:normal}
For each NTL proportion, the feature matrix has at least 20 features, which are the 8 neighborhood features combined with the 12 daily average consumption features. Depending on the distribution of customers in each NTL proportion, up to 16 binary master data features are added.
However, only a fraction of them is expressive enough to improve the prediction results. The number of retained and number of total features per NTL proportion sample are summarized in Table~\ref{table:feature_size}.

\begin{table}[h!]
\renewcommand{\arraystretch}{1.3}
\caption{Number of Features Used Per NTL Proportion}
\label{table:feature_size}
\centering
\begin{tabular}{|c|c|c|}
\hline
NTL prop. & \#Retained binary feat. & \#Total feat. \\
\hline
1\% - 10\% & 5 & 25 \\
\hline
30\% - 70\% & 4 & 24 \\
\hline
20\%, 80\%, 90\% & 6 & 26 \\
\hline
\end{tabular}
\end{table}

In order to optimize the training, each of the 8 neighborhood features and 12 daily average consumption features is normalized:
\begin{align}
x_j \prime = \frac{x_j - \bar{x}_j}{\sigma_j}.
\end{align} 

This normalization makes the values of each future in the data have zero mean and unit variance. This allows to reduce the impact of features with a broad range of values. As an outcome, each feature contributes approximately proportionally to the classification.

\subsection{Models}
In this section, we provide an overview of the models used in this paper and explain how they scale to Big Data sets.

\subsubsection{Logistic Regression}
Logistic regression (LR) is a linear classifier that optimizes a convex cross-entropy loss function during the training of the weights \cite{cox1958regression}. It is related to linear regression, but feeds the continuous output value in the Sigmoid function $\sigma(x) = \frac{1}{1 + \exp(-x)}$ in order to predict a probability of binary class membership.
LR scales to Big Data sets, as the minibatch gradient descent, that is used to optimize the weights, can be parallelized among different cores or nodes.

\subsubsection{k-nearest Neighbors}
$k$-nearest neighbors (KNN) is an instance-based or lazy learning method that does not use weights, as there is no training phase as such \cite{altman1992introduction}. During prediction, the class of an example is determined by selecting the majority class of the $k$ nearest training examples. Defining proximity is subject to the selection of a distance function, of which the most popular ones include Euclidean, Manhattan or cosine.
$k$ is a smoothing parameter. The larger $k$, the smoother the output. Since KNN in an instance-based method, predicting is slow and prediction times grow with $k$.
As the prediction of the class of an example in the test set is independent from the other elements, the predictions can be distributed among different cores or nodes.

\subsubsection{Support Vector Machine}
A support vector machine (SVM) \cite{vapnik1999overview} is a maximum margin classifier, i.e. it creates a maximum separation between classes.
Support vectors hold up the separating hyperplane. In practice, they are just a small fraction of the training examples.
Therefore, a SVM is less prone to overfitting than other classifiers, such as a neural network \cite{cao2003support}.
The training of a SVM can be defined as a Lagrangian dual problem having a convex cost function.
By default, the separating hyperplane is linear.
Training of SVMs using a kernel to map the input to higher dimension is only feasible for a few ten thousand training examples in a realistic amount of time \cite{CC01a}.
Therefore, for Big Data sets only a linear implementation of SVMs is practically usable \cite{scikit-learn}.


\subsubsection{Random Forest}
A random forest is an ensemble estimator that comprises a number of decision trees \cite{ho1995random}. Each tree is trained on a subsample of the data and feature set in order to control overfitting. In the prediction phase, a majority vote is made of the predictions of the individual trees.
Training of the individual trees is independent from each other, so it can be distributed among different cores or nodes.

\section{Evaluation}
\label{chapter:eval}

\subsection{Metric}
The performance measure used in the following experiments is the area under the receiver-operating curve (AUC). It plots the true positive rate or recall against the false positive rate. It is particularly useful for NTL detection, as it allows to handle imbalanced datasets and puts correct and incorrect inspection results in relation to each other.
In many applications, multiple thresholds are used to generate points plotted on a receiver-operating curve. However, the AUC can also be computed for a single point, when connecting it with straight lines to $(0, 0)$ and $(1, 1)$ as shown in \cite{van2003area}:
\begin{align}
\operatorname{AUC} = \frac{\operatorname{Recall} + \operatorname{Specificity}}{2},
\end{align}
where the recall or true positive rate or sensitivity is a measure of the proportion of the true positives found:
\begin{align}
\operatorname{Recall} = \frac{\operatorname{TP}}{\operatorname{TP}+\operatorname{FN}}.
\end{align}
The true negative rate or specificity is a measure of the proportion of the true negatives classified as negative:
\begin{align}
\operatorname{Specificity} = \frac{\operatorname{TN}}{\operatorname{TN} + \operatorname{FP}}.
\end{align}
For a binary classification problem, a AUC score of 0.5 is equivalent to chance and a score of greater than 0.5 is better than chance.

\subsection{Implementation Details}
\label{chapter:imp}
All computations were run on a server with 24 cores and 128 GB of RAM. The entire code was written in Python. The neighborhood features were computed using \texttt{Spark} \cite{zaharia2010spark}. For all experiments, \texttt{scikit-learn} \cite{scikit-learn} was used, which allows to distribute the training and evaluation of each of the four classifiers among all cores.

\subsection{Experimental Setup}
For every NTL proportion, the data set is split into training, validation and test sets with a ratio of 80\%, 10\% and 10\%, respectively.
Each of the four models is trained using 10-fold cross-validation. For each of the four models the trained classifier that performed the best on the validation set in any of the 10 folds is selected and tested on the test set to report the test AUC. This methodology is related to \cite{glauner2016large}.
For each of the four models, the following parameter values were determined empirically as a compromise between expressiveness, generalization of models and training time. For logistic regression and SVM, the inverse regularization factor $C$ is set to 1.0. $K=100$ neighbors are visited in KNN. The random forest consists of 1K trees.

Running all experiments used in this paper including cross-validation takes about 4 hours on the computing infrastructure described in Section~\ref{chapter:imp}.

\subsection{Results}
\begin{figure}[h!]
\centering
\includegraphics[width=0.475\textwidth]{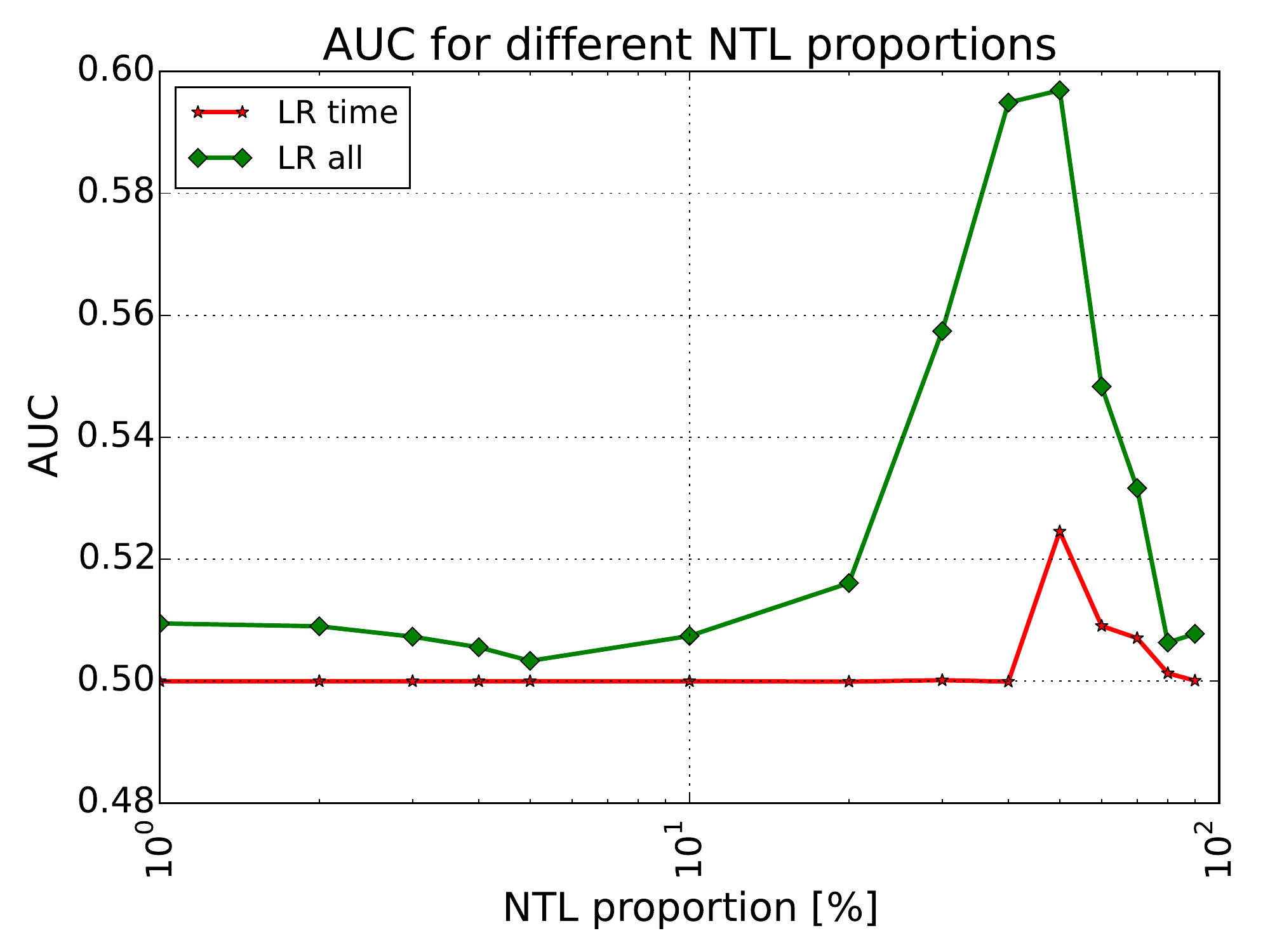}
\caption[XXX]{Test performance of logistic regression classifier on different NTL proportions for time series and all features.}
\label{fig:AUC_LR}
\end{figure}

For different NTL proportions, the test AUC of the logistic regression (LR) classifiers is depicted in Figure~\ref{fig:AUC_LR}. Using only the time series daily average consumption features of the last 12 months results in a classifier that performs like chance for most NTL proportions. It only performs better than chance for NTL proportions of 50\%-80\% with a maximum AUC of 0.525 for a NTL proportion of 50\%. However, by adding the neighborhood and selected categorial features, the classifier performs noticeably better than chance for all NTL proportions and significantly better than time series features only for NTL proportions of 30\%-70\%.

Similar experiments are run for the KNN, SVM and random forest (RF) classifiers and summarized Table~\ref{table:res1}. It can be observed that the extra features help all classifiers to maximize the overall AUC scores and that the classifiers perform noticeably better than chance for more NTL proportions.

\begin{table}[h!]
\setlength{\tabcolsep}{1pt}
\renewcommand{\arraystretch}{1.3}
\caption{Comparison of Classifiers Trained on Time Series and All Features}
\begin{minipage} {0.475\textwidth}
\label{table:res1}
\begin{center}
\begin{tabular}{|c|c|c|c|c|c|c|c|c|}
\hline
NTL & LR$_t$ & LR$_a$ & KNN$_t$ & KNN$_a$ & SVM$_t$ & SVM$_a$ & RF$_t$ & RF$_a$ \\
\hline
1\% & 0.5 & 0.51 & 0.5 & 0.5 & 0.5 & 0.5 & 0.5 & 0.505 \\
\hline
2\% & 0.5 & 0.509 & 0.5 & 0.5 & 0.5 & 0.5 & 0.5 & 0.505 \\
\hline
3\% & 0.5 & 0.507 & 0.5 & 0.5 & 0.5 & 0.505 & 0.5 & 0.511 \\
\hline
4\% & 0.5 & 0.506 & 0.5 & 0.5 & 0.5 & 0.503 & 0.502 & 0.509 \\
\hline
5\% & 0.5 & 0.503 & 0.5 & 0.5 & 0.5 & 0.504 & 0.5 & 0.511 \\
\hline
10\% & 0.5 & 0.507 & 0.504 & 0.5 & 0.5 & 0.505 & 0.504 & 0.519 \\
\hline
20\% & 0.5 & 0.516 & 0.523 & 0.506 & 0.5 & 0.511 & 0.509 & 0.539 \\
\hline
30\% & 0.5 & 0.557 & 0.53 & 0.549 & 0.5 & 0.552 & 0.535 & 0.578 \\
\hline
40\% & 0.5 & 0.595 & 0.546 & 0.587 & 0.5 & 0.592 & 0.55 & \textbf{0.619} \\
\hline
50\% & \textbf{0.525} & \textbf{0.597} & \textbf{0.57} & \textbf{0.596} & \textbf{0.521} & \textbf{0.6} & 0.572 & 0.618 \\
\hline
60\% & 0.509 & 0.548 & 0.545 & 0.556 & 0.509 & 0.546 & \textbf{0.579} & 0.582 \\
\hline
70\% & 0.507 & 0.532 & 0.526 & 0.53 & 0.507 & 0.529 & 0.55 & 0.553 \\
\hline
80\% & 0.501 & 0.506 & 0.508 & 0.505 & 0.502 & 0.51 & 0.527 & 0.514 \\
\hline
90\% & 0.5 & 0.508 & 0.5 & 0.5 & 0.502 & 0.506 & 0.507 & 0.506 \\
\hline
\end{tabular}
\end{center}
Subscript $t$ denotes that only the time series is used in the models.
Subscript $a$ denotes that all features are used: time series, neighborhood features and selected master data.
Best proportion per model in \textbf{bold}.
\end{minipage}
\end{table}

\begin{figure}[h!]
\centering
\includegraphics[width=0.475\textwidth]{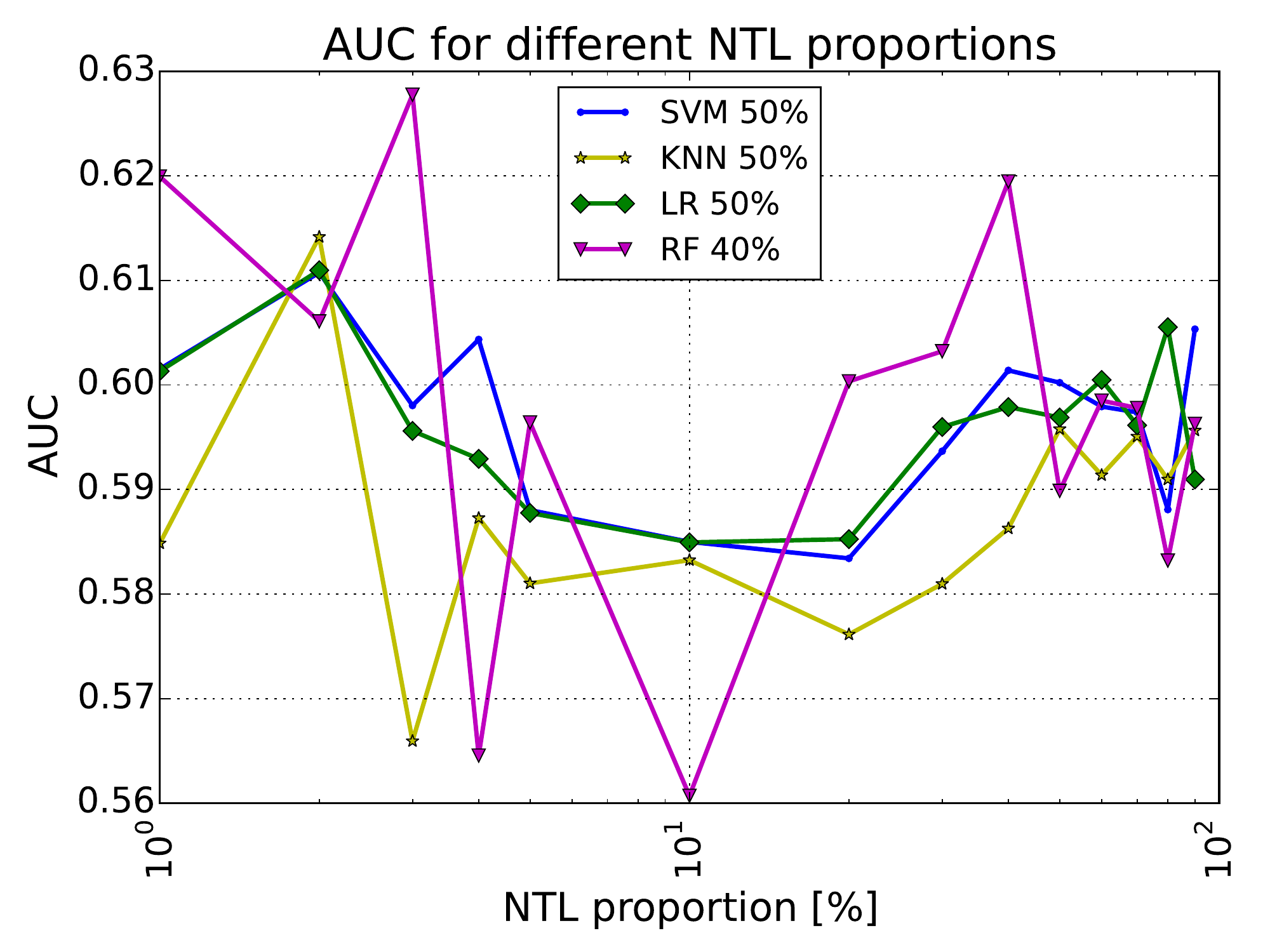}
\caption[XXX]{Test performance of optimized classifiers on different NTL proportions.}
\label{fig:AUC_vs}
\end{figure}

The LR, KNN and SVM classifiers perform the best for a balanced data set of 50\%. The RF classifiers perform the best for 60\% and 40\% using only the time series or all features, respectively. However, it must be noted that for 50\%, both RF classifiers perform close to the optimal AUC scores achieved. This is most likely due to the ensemble, which allows to better adopt to variations in the data set. The four models that performed the best on all features are then tested on all proportions. The results are summarized in Table~\ref{table:AUC_vs} and visualized in Figure~\ref{fig:AUC_vs}.

\begin{table}[h!]
\setlength{\tabcolsep}{2pt}
\renewcommand{\arraystretch}{1.3}
\caption{Performance of Optimized Models on All NTL Proportions}
\begin{minipage} {0.475\textwidth}
\label{table:AUC_vs}
\begin{center}
\begin{tabular}{|c|c|c|c|c|}
\hline
NTL prop. & LR 50\% & KNN 50\% & SVM 50\% & RF 40\% \\
\hline
1\% & 0.601 & 0.585 & 0.602 & \textbf{0.62} \\
\hline
2\% & 0.611 & \textbf{0.614} & 0.611 & 0.606 \\
\hline
3\% & 0.596 & 0.566 & 0.598 & \textbf{0.628} \\
\hline
4\% & 0.593 & 0.587 & \textbf{0.604} & 0.565 \\
\hline
5\% & 0.588 & 0.581 & 0.588 & \textbf{0.596} \\
\hline
10\% & \textbf{0.585} & 0.583 & \textbf{0.585} & 0.561 \\
\hline
20\% & 0.585 & 0.576 & 0.583 & \textbf{0.6} \\
\hline
30\% & 0.596 & 0.581 & 0.594 & \textbf{0.603} \\
\hline
40\% & 0.598 & 0.586 & 0.601 & \textbf{0.619} \\
\hline
50\% & 0.597 & 0.596 & \textbf{0.6} & 0.59 \\
\hline
60\% & \textbf{0.6} & 0.591 & 0.598 & 0.598 \\
\hline
70\% & 0.596 & 0.595 & 0.597 & \textbf{0.598} \\
\hline
80\% & \textbf{0.606} & 0.591 & 0.588 & 0.583 \\
\hline
90\% & 0.591 & 0.596 & \textbf{0.605} & 0.596 \\
\hline
\hline
Max & 0.611 & 0.614 & 0.611 & 0.628 \\
\hline
Min & 0.585 & 0.566 & 0.583 & 0.561 \\
\hline
$\overline{AUC}$ & 0.5959 & 0.5877 & 0.5967 & 0.5973 \\
\hline
$\sigma_{AUC}$ & 0.0071 & 0.0108 & 0.0079 & 0.0183 \\
\hline
\end{tabular}
\end{center}
Model XY\% stands for a model that was trained on a NTL proportion of XY\% and tested on all proportions.
Best model per proportion in \textbf{bold}.
\end{minipage}
\end{table}

The RF classifier achieves the greatest AUC throughout the experiments of 0.628 for a NTL proportion of 3\% and achieves the best AUC among all classifiers for 7 of the 14 classifiers. The SVM performs the best on 4 proportions, the LR performs performs the best on 2 proportions. Both classifiers perform similarly well on the NTL proportion of 10\%. The KNN classifier only performs the best on one proportion. Even though the RF achieved the maximum AUC, it also has the lowest AUC throughout the experiments. Furthermore, it has the greatest standard deviation of all classifiers.

\subsection{Discussion}
Overall, all four classifiers perform in the same regime, as their mean AUC scores over all NTL proportions are very close. This observation is often made in machine learning, as the actual algorithm is less important, but having more and representative data is generally considered to be more important \cite{banko2001scaling}. This can also be justified by the "no free lunch theorem", which states that no learning algorithm is generally better than others \cite{wolpert1996lack}.
In our previous work, we only used the features derived from the consumption time series \cite{glauner2016large}. Using also the neighborhood information and categorial customer master data, each of the four classifiers consistently performs better than the classifiers in our previous work for all NTL proportions.

In Section~\ref{chapter:review}, we have identified the main challenges to advance NTL detection. We believe that NTL detection using the current inspection labels is limited. Therefore, it is desirable to analyze the data in an unsupervised manner in order to get further insights into its structure. This will help to compensate the bias in the distribution of the labels and to also remove potentially wrong inspection labels.

\section{Conclusion and Future Work}
\label{chapter:end}
In this work, we have proposed two neighborhood features for NTL detection of a Big Data set of 700K customers and 400K inspection results by splitting the area into a grid: the ratio of customers inspected and ratio of inspected customers for which NTL was detected. We generated these features for four different grid sizes. We have analyzed the statistical properties of their distributions and showed why they are useful for predicting NTL.
These features were combined with daily average consumption features of the last 12 months before the most recent inspection of a customer from a Big Data set, which contains 32M meter readings in total. Furthermore, we also used selected customer master data, such as the customer class and voltage of the connection of the customer.
We used four machine learning algorithms that are particularly suitable for Big Data sets to predict if a customer causes a NTL or not: logistic regression, k-nearest neighbors, linear support vector machine and random forest.
We observed that all models significantly perform better when using the neighborhood and customer master data features compared to using only the time series features. All models perform in the same regime measured by the AUC score. In total, the random forest classifier slightly outperforms the other classifiers.

In our future research, we are planning to investigate the bias in the inspection labels and how to correct it using unsupervised methods. We believe that bias-free samples of data will allow to train more general and accurate NTL detection models.

\bibliographystyle{abbrv}
\bibliography{sigproc}  

\begin{thebibliography}{10}

\bibitem{alam2004power}
M.~Alam, E.~Kabir, M.~Rahman, and M.~Chowdhury.
\newblock Power sector reform in bangladesh: Electricity distribution system.
\newblock {\em Energy}, 29(11):1773--1783, 2004.

\bibitem{altman1992introduction}
N.~S. Altman.
\newblock An introduction to kernel and nearest-neighbor nonparametric
  regression.
\newblock {\em The American Statistician}, 46(3):175--185, 1992.

\bibitem{angelos2011detection}
E.~W.~S. Angelos, O.~R. Saavedra, O.~A.~C. Cort{\'e}s, and A.~N. de~Souza.
\newblock Detection and identification of abnormalities in customer
  consumptions in power distribution systems.
\newblock {\em IEEE Transactions on Power Delivery}, 26(4):2436--2442, 2011.

\bibitem{banko2001scaling}
M.~Banko and E.~Brill.
\newblock Scaling to very very large corpora for natural language
  disambiguation.
\newblock In {\em Proceedings of the 39th annual meeting on association for
  computational linguistics}, pages 26--33. Association for Computational
  Linguistics, 2001.

\bibitem{cao2003support}
L.-J. Cao and F.~E.~H. Tay.
\newblock Support vector machine with adaptive parameters in financial time
  series forecasting.
\newblock {\em IEEE Transactions on neural networks}, 14(6):1506--1518, 2003.

\bibitem{CC01a}
C.-C. Chang and C.-J. Lin.
\newblock {LIBSVM}: A library for support vector machines.
\newblock {\em ACM Transactions on Intelligent Systems and Technology},
  2:27:1--27:27, 2011.
\newblock Software available at \url{http://www.csie.ntu.edu.tw/~cjlin/libsvm}.

\bibitem{chauhan2013non}
A.~Chauhan and S.~Rajvanshi.
\newblock Non-technical losses in power system: A review.
\newblock In {\em Power, Energy and Control (ICPEC), 2013 International
  Conference on}, pages 558--561. IEEE, 2013.

\bibitem{cox1958regression}
D.~R. Cox.
\newblock The regression analysis of binary sequences.
\newblock {\em Journal of the Royal Statistical Society. Series B
  (Methodological)}, pages 215--242, 1958.

\bibitem{decarlo1997meaning}
L.~T. DeCarlo.
\newblock On the meaning and use of kurtosis.
\newblock {\em Psychological methods}, 2(3):292, 1997.

\bibitem{depuru2013high}
S.~S. S.~R. Depuru, L.~Wang, V.~Devabhaktuni, and R.~C. Green.
\newblock High performance computing for detection of electricity theft.
\newblock {\em International Journal of Electrical Power \& Energy Systems},
  47:21--30, 2013.

\bibitem{di2012improving}
M.~Di~Martino, F.~Decia, J.~Molinelli, and A.~Fern{\'a}ndez.
\newblock Improving electric fraud detection using class imbalance strategies.
\newblock In {\em ICPRAM (2)}, pages 135--141, 2012.

\bibitem{glauner2016challenge}
P.~Glauner, A.~Boechat, L.~Dolberg, J.~Meira, R.~State, F.~Bettinger,
  Y.~Rangoni, and D.~Duarte.
\newblock The challenge of non-technical loss detection using artificial
  intelligence: A survey.
\newblock {\em arXiv preprint arXiv:1606.00626}, 2016.

\bibitem{glauner2016large}
P.~Glauner, A.~Boechat, L.~Dolberg, R.~State, F.~Bettinger, Y.~Rangoni, and
  D.~Duarte.
\newblock Large-scale detection of non-technical losses in imbalanced data
  sets.
\newblock In {\em Innovative Smart Grid Technologies Conference (ISGT), 2016
  IEEE Power \& Energy Society}. IEEE, 2016.

\bibitem{ho1995random}
T.~K. Ho.
\newblock Random decision forests.
\newblock In {\em Document Analysis and Recognition, 1995., Proceedings of the
  Third International Conference on}, volume~1, pages 278--282. IEEE, 1995.

\bibitem{jiang2014energy}
R.~Jiang, R.~Lu, Y.~Wang, J.~Luo, C.~Shen, and X.~S. Shen.
\newblock Energy-theft detection issues for advanced metering infrastructure in
  smart grid.
\newblock {\em Tsinghua Science and Technology}, 19(2):105--120, 2014.

\bibitem{kazerooni2014literature}
M.~Kazerooni, H.~Zhu, and T.~J. Overbye.
\newblock Literature review on the applications of data mining in power
  systems.
\newblock In {\em Power and Energy Conference at Illinois (PECI), 2014}, pages
  1--8. IEEE, 2014.

\bibitem{mclaughlin2009energy}
S.~McLaughlin, D.~Podkuiko, and P.~McDaniel.
\newblock Energy theft in the advanced metering infrastructure.
\newblock In {\em International Workshop on Critical Information
  Infrastructures Security}, pages 176--187. Springer, 2009.

\bibitem{nagi2010nontechnical}
J.~Nagi, K.~S. Yap, S.~K. Tiong, S.~K. Ahmed, and M.~Mohamad.
\newblock Nontechnical loss detection for metered customers in power utility
  using support vector machines.
\newblock {\em IEEE transactions on Power Delivery}, 25(2):1162--1171, 2010.

\bibitem{scikit-learn}
F.~Pedregosa, G.~Varoquaux, A.~Gramfort, V.~Michel, B.~Thirion, O.~Grisel,
  M.~Blondel, P.~Prettenhofer, R.~Weiss, V.~Dubourg, J.~Vanderplas, A.~Passos,
  D.~Cournapeau, M.~Brucher, M.~Perrot, and E.~Duchesnay.
\newblock Scikit-learn: Machine learning in {P}ython.
\newblock {\em Journal of Machine Learning Research}, 12:2825--2830, 2011.

\bibitem{ramos2012identification}
C.~C.~O. Ramos, A.~N. De~Souza, D.~S. Gastaldello, and J.~P. Papa.
\newblock Identification and feature selection of non-technical losses for
  industrial consumers using the software weka.
\newblock In {\em Industry Applications (INDUSCON), 2012 10th IEEE/IAS
  International Conference on}, pages 1--6. IEEE, 2012.

\bibitem{smith2004electricity}
T.~B. Smith.
\newblock Electricity theft: a comparative analysis.
\newblock {\em Energy Policy}, 32(18):2067--2076, 2004.

\bibitem{van2003area}
W.~B. van~den Hout.
\newblock The area under an roc curve with limited information.
\newblock {\em Medical decision making}, 23(2):160--166, 2003.

\bibitem{vapnik1999overview}
V.~N. Vapnik.
\newblock An overview of statistical learning theory.
\newblock {\em IEEE transactions on neural networks}, 10(5):988--999, 1999.

\bibitem{wolpert1996lack}
D.~H. Wolpert.
\newblock The lack of a priori distinctions between learning algorithms.
\newblock {\em Neural computation}, 8(7):1341--1390, 1996.

\bibitem{zaharia2010spark}
M.~Zaharia, M.~Chowdhury, M.~J. Franklin, S.~Shenker, and I.~Stoica.
\newblock Spark: cluster computing with working sets.
\newblock {\em HotCloud}, 10:10--10, 2010.

\end{thebibliography}

\end{document}